\documentclass[11pt,a4paper]{article}
\usepackage[hyperref]{acl2019}

\usepackage[utf8]{inputenc} 
\usepackage[T1]{fontenc}    
\usepackage{url}            
\usepackage{booktabs}       
\usepackage{amsfonts}       
\usepackage{nicefrac}       
\usepackage{microtype}      
\usepackage{bm}

\usepackage{times}
\usepackage{latexsym}
\usepackage{helvet}  
\usepackage{courier}  
\usepackage{graphicx}
\usepackage{subfigure}  

\usepackage{color}
\usepackage{multirow}
\usepackage{bbm}
\usepackage{algorithm,algorithmic}
\usepackage{float}
\usepackage{caption}
\usepackage{wrapfig}
\usepackage{amsmath}
\usepackage{lipsum}

\newcommand{\Imat}{{\bf I}}

\newcommand{\Wmat}[0]{{{\bf W}}}

\newcommand{\cv}[0]{{\boldsymbol{c}}}

\newcommand{\ev}[0]{{\boldsymbol{e}}}

\newcommand{\hv}[0]{{\boldsymbol{h}}}

\newcommand{\vv}{\boldsymbol{v}}

\newcommand{\xv}{\boldsymbol{x}}
\newcommand{\yv}{\boldsymbol{y}}
\newcommand{\zv}{\boldsymbol{z}}

\newcommand{\Sigmamat}[0]{{\boldsymbol{\Sigma}}}

\newcommand{\muv}[0]{{\boldsymbol{\mu}}}

\newcommand{\sigmav}[0]{{\boldsymbol{\sigma}}}

\newcommand{\Ncal}{\mathcal{N}}

\aclfinalcopy 
\def\aclpaperid{1593} 

\title{Syntax-Infused Variational Autoencoder for Text Generation}

\author{Xinyuan Zhang$^1$\thanks{\; Part of this work was done when the first two authors were at Bloomberg.}, Yi Yang$^2$\footnotemark[1], Siyang Yuan$^1$, Dinghan Shen$^1$, Lawrence Carin$^1$\\
  $^1$Duke University\\
  $^2$ASAPP Inc.\\
  \texttt{xy.zhang@duke.edu}, \texttt{yyang@asapp.com}}
\date{}

\begin{document}
\maketitle

\begin{abstract}
We present a syntax-infused variational autoencoder (SIVAE), that integrates sentences with their syntactic trees to improve the grammar of generated sentences.
Distinct from existing VAE-based text generative models, SIVAE contains two separate latent spaces, for sentences and syntactic trees.
The evidence lower bound objective is redesigned correspondingly, by optimizing a joint distribution that accommodates two encoders and two decoders.
SIVAE works with long short-term memory architectures to simultaneously generate sentences and syntactic trees.
Two versions of SIVAE are proposed: one captures the dependencies between the latent variables through a conditional prior network, and the other treats the latent variables independently such that syntactically-controlled sentence generation can be performed.
Experimental results demonstrate the generative superiority of SIVAE on both reconstruction and targeted syntactic evaluations.
Finally, we show that the proposed models can be used for unsupervised paraphrasing given different syntactic tree templates.
\end{abstract}

\section{Introduction}\label{sec:intro}
Neural language models based on recurrent neural networks \cite{mikolov2010recurrent} and sequence-to-sequence architectures \cite{sutskever2014sequence} have revolutionized the NLP world.
Deep latent variable modes, in particular, the variational autoencoders  (VAE)~\cite{kingma2014auto,rezende2014stochastic} integrating inference models with neural language models have been widely adopted on text generation~\cite{bowman2016generating,yang2017improved,kim2018semi}, where the encoder and  the decoder are modeled by long short-term memory (LSTM) networks \cite{chung2014empirical}.
For a random vector from the latent space representing an unseen input, the decoder can generate realistic-looking novel data in the context of a text model, making the VAE an attractive generative model.
Compared to simple neural language models, the latent representation in a VAE is supposed to give the model more expressive capacity.

\begin{figure}
	\centering
	\includegraphics[width=3in]{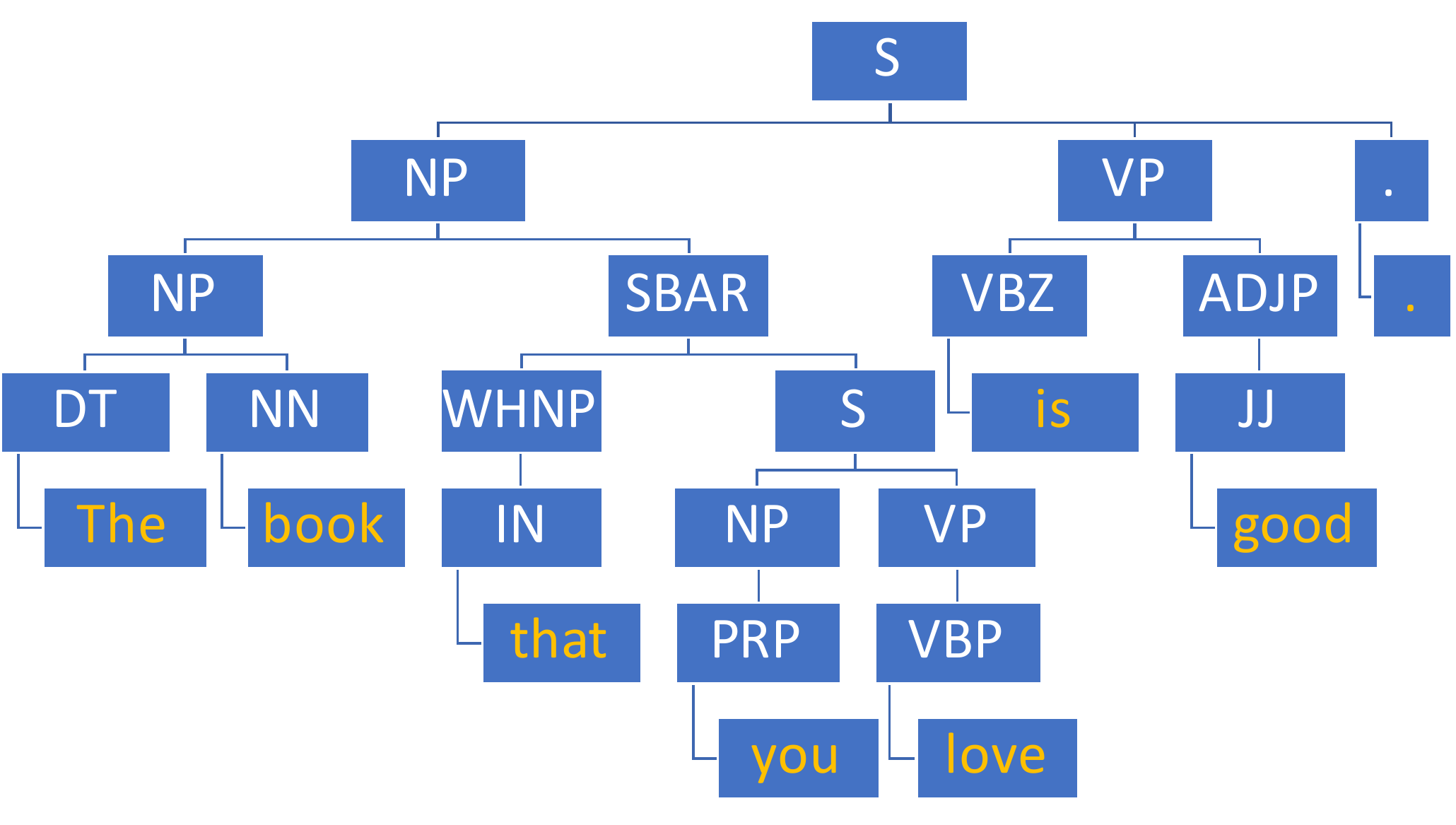}
	\caption{An example of a constituency tree structure.}
	\label{fig:tree}
\end{figure}

Although syntactic properties can be implicitly discovered by such generative models, \citeauthor{shi2016does} \shortcite{shi2016does} show that many deep structural details are still missing in the generated text.
As a result of the absence of explicit syntactic information, generative models often produce ungrammatical sentences.
To address this problem, recent works attempt to leverage explicit syntactic knowledge to improve the quality of machine translation \cite{eriguchi2016tree,bastings2017graph,chen2017improved} and achieve good results.
Motivated by such success, we suggest that deep latent variable models for text generation can also benefit from the incorporation of syntactic knowledge.
Instead of solely modeling sentences, we want to utilize augmented data by introducing an auxiliary input, a syntactic tree, to enrich the latent representation and make the generated sentences more grammatical and fluent.
Syntactic trees can either be obtained from existing human-labeled trees or syntactically parsed sentences using well-developed parsers.
An example of a constituency tree is shown in Figure \ref{fig:tree}.
In this work, we remove leaf nodes and linearize the bracketed parse structure into a syntactic tree sequence to simplify the encoding and decoding processes.
For example, the syntactic tree sequence for the sentence ``The book that you love is good.'' is \texttt{\small(S(NP(NP(DT)(NN))(SBAR(WHNP(IN))(S(NP(PRP ))(VP(VBP)))))(VP(VBZ)(ADJP(JJ)))(.))}.
Given such data, we aim to train a latent variable model that jointly encodes and decodes a sentence and its syntactic tree.


We propose a syntax-infused VAE model to help improve generation, by integrating syntactic trees with sentences.
In contrast to the current VAE-based sentence-generation models, a key differentiating aspect of SIVAE is that we map the sentences and the syntactic trees into two latent representations, and generate them separately from the two latent spaces.
This design decouples the semantic and syntactic representations and makes it possible to concentrate generation with respect to either syntactic variation or semantic richness.
To accommodate the two latent spaces in one VAE framework, the evidence lower bound (ELBO) objective needs to be redesigned based on optimizing the joint log likelihood of sentences and syntactic trees.
This new objective makes SIVAE a task-agnostic model, with two encoders and two decoders, so that it can be further used for other generative tasks.

Two variants of SIVAE that differ in the forms of the prior distributions corresponding to the syntactic tree latent variables are presented.
SIVAE-c captures dependencies between two latent variables by making the syntax prior conditioned on the sentence prior.
During generation, we first sample a latent variable from the sentence latent space and then sample the syntactic tree latent variable depending on the sampled sentence latent variable.
This process resembles how humans write: think about substances like entities and topics first, then realize with a specific syntactic structure.
We further propose SIVAE-i assuming the two priors are independent, and change the ELBO of the joint log likelihood correspondingly.
This independence assumption manifests syntactically-controlled sentence generation as it allows to alter the syntactic structure, desirable for related tasks like paraphrase generation.
Given a sentence and a syntactic tree template, the model produces a paraphrase of the sentence whose syntax conforms to the template.
Our SIVAE-based paraphrasing network is purely unsupervised, which makes it particularly suitable for generating paraphrases in low-resource languages or types of content.

The experiments are conducted on two datasets: one has trees labeled by humans and the other has trees parsed by a state-of-the-art parser \citep{kitaev2018constituency}.
Other than employing the standard language modeling evaluation metrics like perplexity, we also adopt the targeted syntactic evaluation \cite{marvin2018targeted} to verify whether the incorporation of syntactic trees improves the grammar of generated sentences.
Experiments demonstrate that the proposed model improves the quality of generated sentences compared to other baseline methods, on both the reconstruction and grammar evaluations.
The proposed methods show the ability for unsupervised paraphrase generation under different syntactic tree templates.

Our contributions are four-fold:
$i$) We propose a syntax-infused VAE that integrates syntactic trees with sentences, to grammatically improve the generated sentences.
$ii$) We redesign the ELBO of the joint log likelihood, to accommodate two separate latent spaces in one VAE framework, for two SIVAE model variants based on different intuitions, which can be further used for other applications.
$iii$) We evaluate our models on data with human-constituted trees or parsed trees, and yield promising results in generating sentences with better reconstruction loss and less grammatical errors, compared to other baseline methods.
$iv$) We present an unsupervised paraphrasing network based on SIVAE-i that can perform syntactically controlled paraphrase generation.

\begin{figure*}
	\centering
	\includegraphics[width=6in]{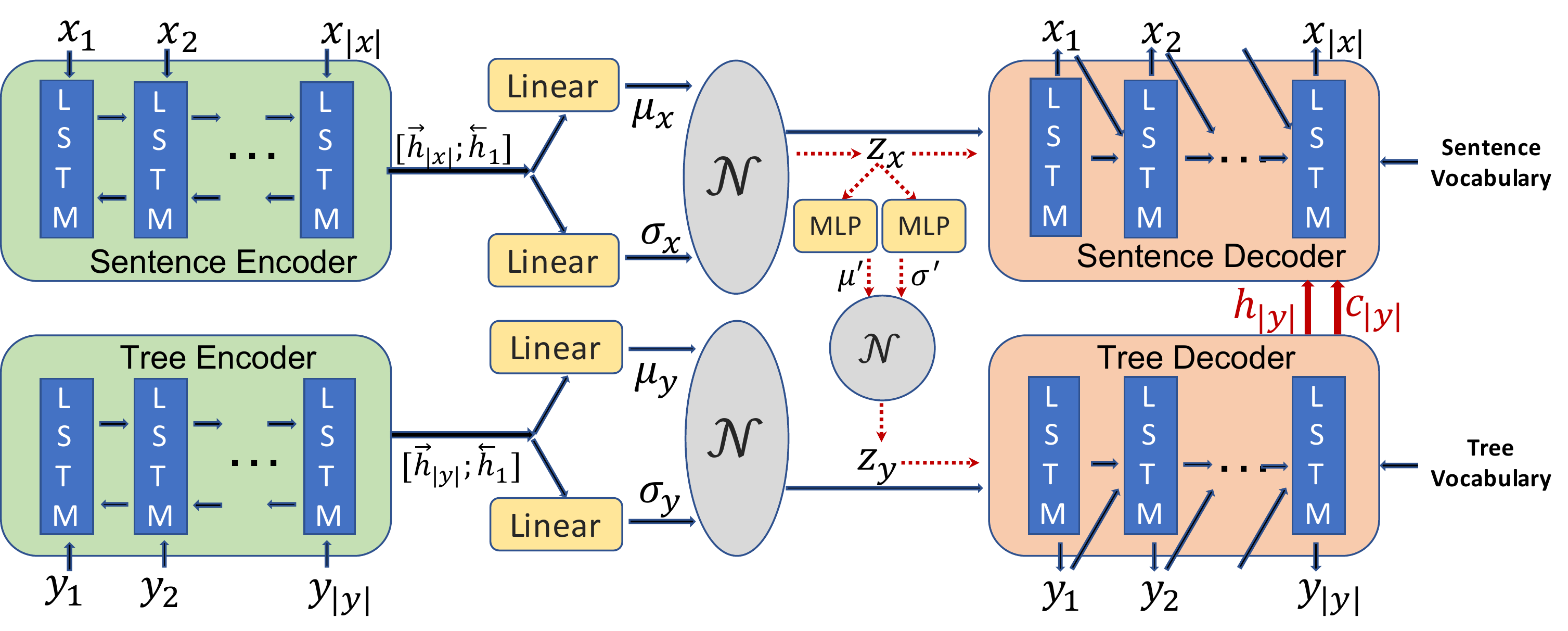}
	\caption{Block diagram of the proposed SIVAE model encoding and decoding sentences and their syntactic trees jointly. The prior network (dashed lines) is used only for the sampling stage of SIVAE-c.}
	\label{fig:diag}
\end{figure*}

\section{Methodology}\label{sec:model}
Given a sentence $\xv$ and its corresponding syntactic tree $\yv$, the goal is to jointly encode $\xv$ and $\yv$ into latent representations $\zv_x\in\mathbb{R}^d$ and $\zv_y\in\mathbb{R}^d$, and then decode them jointly from the two latent spaces.
We employ the VAE framework such that realistic-looking novel sentences can be generated with randomly sampled latent representations.
However, current VAE-based language models cannot accommodate two separate latent spaces for $\zv_x$ and $\zv_y$.
To incorporate $\xv$, $\yv$, $\zv_x$, and $\zv_y$ in one VAE framework, the objective needs to be redesigned to optimize the log joint likelihood $\log p(\xv,\yv)$.
We propose two model variants of SIVAE.
The first model (SIVAE-c; Section \ref{sec:p2}), directly capturing the dependencies between $\zv_x$ and $\zv_y$, presumes that semantic information should influence syntax structure.
During the sampling stage, the prior for $\zv_y$ is drawn based on $\zv_x$ from a conditional prior network $p(\zv_y|\zv_x)$; $\zv_x$ implicitly encodes the subject of the sentence, and $\zv_y$ encodes the corresponding syntax.
Although this model has robust performance on generation, it doesn't allow us to syntactically control the generated sentences by freely changing the syntactic tree template in $\zv_y$.
Thus we propose SIVAE-i (Section \ref{sec:p1}), which generates sentences and syntactic trees assuming the priors $p(\zv_x)$ and $p(\zv_y)$ are independent.
The entire architecture is shown in Figure \ref{fig:diag}.
\subsection{Modeling Syntax-Semantics Dependencies}\label{sec:p2}
Since the syntax of a sentence is influenced by the semantics, especially when the content is long, we first propose a generative model to exploit the dependencies between $\zv_x$ and $\zv_y$, through a conditional prior network $p_\psi(\zv_y|\zv_x)$.
Formally, SIVAE-c models the joint probability of the sentence and its syntactic tree:
\begin{align}
\nonumber p(\xv,\yv)=\int_{d_{\zv_x}}\int_{d_{\zv_y}}&p(\xv|\yv,\zv_x)p(\yv|\zv_x,\zv_y)\cdot\\ &p(\zv_y|\zv_x)p(\zv_x)d_{\zv_y}d_{\zv_x},
\end{align}
where the prior over $\zv_x$ is the isotropic Gaussian $p(\zv_x) = \Ncal(\mathbf{0},\Imat)$.
We define $q(\cdot)$ to be the variational posterior distributions that approximate the true posterior distributions.
The model is trained by maximizing the lower bound of the log likelihood
\begin{align}\label{eq:p2}
&\log p (\xv, \yv)\geq\mathcal{L}(\xv,\yv;\theta,\phi,\psi)=\\
\nonumber&\mathbb{E}_{q_\phi(\zv_x|\xv)} \log p_\theta(\xv|\yv,\zv_x) - \operatorname{KL} [q_\phi(\zv_x | \xv)||p(\zv_x)]\\
\nonumber&+\mathbb{E}_{q_\phi(\zv_y|\yv,\zv_x)} \log p_\theta(\yv | \zv_y)\\
\nonumber&-\operatorname{KL} [ q_\phi(\zv_y | \yv, \zv_x) || p_\psi(\zv_y | \zv_x) ],
\end{align}
where $\psi$, $\phi$, and $\theta$ are the parameters of the prior network, the recognition networks, and the generation networks, respectively.
We apply the reparameterize trick to yield a differentiable unbiased estimator of the lower bound objective.

\paragraph{Conditional Prior Network}
The key to SIVAE-c is the conditional prior which is used to model the dependencies between the sentence latent variable $\zv_x$ and the syntactic tree latent variable $\zv_y$.
Given $\zv_x$, the prior for $\zv_y$ is sampled from a conditional probability $p_\psi(\zv_y|\zv_x)$ modeled by a multivariate Gaussians $\mathcal{N}(\muv^\prime,\sigmav^{\prime 2}\Imat)$.
The parameters of the Gaussian distribution are computed from $\zv_x$ with a conditional prior network parameterized by $\psi$. 
In particular, $\muv^\prime$ and $\sigmav^{\prime 2}$ are the outputs of multilayer perceptron (MLP) networks taking $\zv_x$ as the input.

\paragraph{Recognition Networks}
To differentiate through the sampling stage $\zv\sim q_\phi(\zv|\xv)$, the VAE encoder $q_\phi(\zv_x|\xv)$ is also assumed to be a Gaussian distribution $\mathcal{N}(\muv_x,\Sigmamat_x)$, where $\muv(\xv)$ and $\operatorname{diag}(\Sigmamat(\xv))$ are the outputs of feedforward networks taking $\xv$ as the input.
The recognition network consists of a bidirectional LSTM encoder to produce a sentence embedding for $\xv$ and two linear networks to transform the embedding to the Gaussian parameters.
The Kullback-Leibler (KL) divergence between $q_\phi(\zv_x|\xv)$ and the isotropic Gaussian prior $p(\zv_x)$ is
\begin{align}\label{eq:kl1}
\nonumber&\operatorname{KL}(q_\phi(\zv_x|\xv)\|p(\zv_x))=\frac{1}{2}[-\log|\Sigmamat_x|\\
&-d+\operatorname{tr}(\Sigmamat_x)+\muv_x^T\muv_x].
\end{align}
So we only need to model $\muv_x$ and the diagonal of $\Sigmamat_x$ to compute the KL divergence.

To reconcile the conditional prior $p_\psi(\zv_y|\zv_x)$, the variational posterior $q_\phi(\zv_y|\yv,\zv_x)=\mathcal{N}(\muv_y,\sigmav_y^{2}\Imat)$, also depends on the latent variable $\zv_x$. 
$\muv_y$ and $\sigmav_y^{2}$ are obtained from a recognition network that contains a bidirectional LSTM encoder, producing a syntactic tree embedding, and two linear networks, taking the embedding and $\zv_x$ as inputs.
The KL divergence is then given by
\begin{align}
\nonumber&\operatorname{KL}(q_\phi(\zv_y|\yv,\zv_x)\|p_\psi(\zv_y|\zv_x))=\\
\nonumber&\frac{1}{2}[\log|\sigmav^{\prime 2}\Imat|-\log|\sigmav_y^2\Imat|-d+\operatorname{tr}(\frac{\sigmav_y^2\Imat}{\sigmav^{\prime 2}\Imat})\\
&+(\muv^\prime-\muv_y)^T\sigmav^{\prime-2}\Imat(\muv^\prime-\muv_y)].
\end{align}

\paragraph{Generation Networks}
We employ an LSTM to generate $\yv$ from $p_\theta (\yv | \zv_y)$. A word $v_y$ is selected by computing the probability of $\yv_t=v_y$ conditioned on previously generated words $\yv_{-t}$ and $\zv_y$
\begin{align}
p(\yv_t=v_y|\yv_{-t},\zv_y)\propto\exp((\vv_y^T\Wmat^y\hv_t^y)),
\end{align}
where $\hv_t^y$ is the current hidden states of the LSTM tree decoder
\begin{align}\label{eq:ylstm}
\hv_t^y=\operatorname{LSTM}(\zv_y,\ev(y_{t-1}),\hv_{t-1}^y,\cv_{t-1}^y).
\end{align}
To generate $\xv$ from $p_\theta (\xv | \yv, \zv_x)$, we modify the generative model in GNMT \cite{shah2018generative}.
First, the last hidden states $\hv_{|y|}^y$ and $\cv_{|y|}^y$ in (\ref{eq:ylstm}) are directly used as the generated syntactic tree $\yv$, where $|y|$ is the length of $\yv$.
Then we use another LSTM for sentence generation,
\begin{align}
\hv_t^x=\operatorname{LSTM}(\zv_x,\ev(x_{t-1}),\hv_{|y|}^y,\hv_{t-1}^x,\cv_{t-1}^x).
\end{align}
The conditional probabilities of $\xv_t=v_x$ for $t=1,\cdots,|x|$ are computed as
\begin{align}
p(\xv_t=v_x|\xv_{-t},\zv_x,\yv)\propto\exp((\vv_x^T\Wmat^x\hv_t^x)).
\end{align}
In this way, the generated sentence is conditioned on $\zv_x$ and the generated syntactic tree $\yv$.
SIVAE-c selects possible syntactic tree templates for a given sentence latent variable, but the syntactic tree template cannot be freely determined.

\subsection{Syntactically-Controlled Sentence Generation}\label{sec:p1}
In order to freely change the syntactic tree template embedded in $\zv_y$, we propose an alternative model assuming the independence of two priors.
Let priors $\zv_x$ and $\zv_y$ be independent random variables drawn from $\Ncal(0,\Imat)$.
The variational posteriors $q_\phi(\zv_x | \xv)$ and $q_\phi(\zv_y | \yv)$ follow Gaussian distributions parameterized by the outputs of feedforward networks, whose inputs are $\xv$ and $\yv$.
The model is trained by maximizing the lower bound objective
\begin{align}\label{eq:p1}
&\log p (\xv, \yv)\geq \mathcal{L}(\xv,\yv;\theta,\phi)=\\
\nonumber&\mathbb{E}_{q_\phi (\zv_x|\xv)} \log p_\theta (\xv | \yv, \zv_x) - \operatorname{KL} [q_\phi(\zv_x | \xv) \| p (\zv_x)]\\
\nonumber&+ \mathbb{E}_{q_\phi (\zv_y|\yv)} \log p_\theta (\yv | \zv_y) - \operatorname{KL} [q_\phi(\zv_y | \yv) \| p (\zv_y)].
\end{align}
Since $\yv$ and $\zv_x$ are assumed to be independent when computing the joint probability $p(\xv,\yv)$, we seek to minimize the mutual information $\mathbb{I}(\yv;\zv_x)$ during training.

The recognition networks and the generation networks of SIVAE-i  are similar to those adopted in SIVAE-c, so we omit them for brevity.

\begin{table*}[t!]
	\centering  
	\small
	\begin{tabular}{l|cccccccccc} \toprule
		Dataset & Train & Test & Valid & Ave\_s & Max\_s & Voc\_s & Tree Type & Ave\_t & Max\_t & Voc\_t \\ 
		\midrule
		PTB & 39366 & 4921 & 4921 & 25 & 271 & 24699 & Golden & 113 & 1051 & 1272 \\
		wiki90M & 71952 & 8995 & 8994 & 28 & 318 & 28907 & Parsed & 119 & 1163 & 387 \\\bottomrule
	\end{tabular}
	\caption{Statistics of the two datasets used in this paper. Ave\_s/ Ave\_t, Max\_s/ Max\_t, and Voc\_s/ Voc\_t denote the average length, maximum length, and vocabulary size for sentences/ tree sequences correspondingly.}\label{tab:data}
\end{table*}

\section{Unsupervised Paraphrasing}\label{sec:para}
Paraphrases are sentences with the same meaning but different syntactic structures.
SIVAE allows us to execute syntax transformation, producing the desired paraphrases with variable syntactic tree templates.
The syntactically controlled paraphrase generation is inspired by \newcite{iyyer2018adversarial}; the difference is that our SIVAE-based syntactic paraphrase network is purely unsupervised.
Unsupervised paraphrasing can be performed using both SIVAE-c and SIVAE-i.

One way to generate paraphrases is to perform syntactically controlled paraphrase generation using SIVAE-i.
The latent representations of an input sentence $\zv_x$ and a syntactic tree template $\zv_y$ are fed into SIVAE-i, and the syntax of the generated sentence conforms with the explicitly selected target template.
However, linearized syntactic sequences are relatively long (as shown in Table \ref{tab:data}) and long templates are more likely to mismatch particular input sentences, which may result in nonsensical paraphrase outputs.
Therefore, we use simplified syntactic sequences as templates, by taking the top two levels of the linearized constituency trees.

The paraphrase generative process is:
\begin{enumerate}
	\item Encode the original sentence to $\zv_x$;
	\item Select and encode a syntactic template into $\zv_y$;
	\item Generate the reconstructed syntactic sequence $\yv$ from $p(\yv | \zv_y)$;
	\item Generate the paraphrase of the original sentence that conforms to $\yv$ from $p(\xv | \yv, \zv_x)$.
\end{enumerate}

We can also use a trained SIVAE-c to generate paraphrases.
The paraphrase generation process is similar to sampling from a standard VAE with various tempera.
The difference is that SIVAE-c first selects possible syntactic tree templates using the conditional prior network $p_\psi(\zv_y|\zv_x)$ then generates paraphrases based on the syntactic template and the latent variable.

\section{Related Work}\label{sec:related}
\paragraph{Syntax-Aware Neural Text Generation}
The ability to generate sentences is core to many NLP tasks, such as machine translation \cite{bahdanau2015neural}, summarization \cite{rush2015neural}, and dialogue generation \cite{vinyals2015neural}.
Recent works have shown that neural text generation can benefit from the incorporation of syntactic knowledge \cite{shen2017neural,choe2016parsing}. 
\citeauthor{sennrich2016linguistic} \shortcite{sennrich2016linguistic} propose to augment each source word representation with its corresponding part-of-speech tag, lemmatized form and dependency label; \citeauthor{eriguchi2016tree} \shortcite{eriguchi2016tree} and~\citeauthor{bastings2017graph} \shortcite{bastings2017graph} utilize a tree-based encoder and a graph convolutional network encoder respectively to embed the syntactic parse trees as part of the source sentence representations;
\citeauthor{chen2017improved} \shortcite{chen2017improved} model source-side syntactic trees with a bidirectional tree encoder and tree-coverage decoder;
\citeauthor{eriguchi2017learning} \shortcite{eriguchi2017learning} implicitly leverage linguistic prior by treating syntactic parsing as an auxiliary task.
However, most of these syntax-aware generation works only focus on neural machine translation.

\paragraph{Deep Latent Variable Models}
Deep latent variable models that combine the complementary strengths of latent variable models and deep learning have drawn much attention recently. 
Generative adversarial networks \cite{goodfellow2014generative} and variational autoencoders \cite{kingma2014auto} are the two families of deep generative models that are widely adopted in applications. 
As VAEs allow discrete generation from a continuous space, they have been a popular variant for NLP tasks including text generation~\cite{bowman2016generating,yang2017improved,xu2018spherical,shen2019towards,wang2019topic}.
The flexibility of VAEs also enables adding conditions during inference to perform controlled language generation~\cite{hu2017toward,zhao2017learning}.
Divergent from these VAE-based text generation models, our work decouples the latent representations corresponding to the sentence and its syntactic tree respectively.

\paragraph{Paraphrase Generation}
Due to the similarity between two tasks, neural machine-translation-based models can often be utilized to achieve paraphrase generation \cite{hasan2016neural,mallinson2017paraphrasing}.
Recently, \citeauthor{iyyer2018adversarial} \shortcite{iyyer2018adversarial} proposed to syntactically control the generated paraphrase and \citeauthor{gupta2018deep} \shortcite{gupta2018deep} generate paraphrases in a deep generative architecture.
However, all these methods assume the existence of some parallel paraphrase corpora while unsupervised paraphrase generation has been little explored.

\section{Experiments}\label{sec:exp}

\begin{table*}[t!]
	\centering
	\scriptsize
	\begin{tabular}{c|ccc|ccc|ccc|ccc}
		\toprule
		\multirow{3}{*}{Model} &
		\multicolumn{6}{c|}{PTB} &
		\multicolumn{6}{c}{wiki90M}  \\
		&\multicolumn{3}{c|}{Standard} & \multicolumn{3}{c|}{Inputless}  &\multicolumn{3}{c|}{Standard} & \multicolumn{3}{c}{Inputless}\\
		& PPL & NLL & KL & PPL & NLL & KL & PPL & NLL & KL & PPL & NLL & KL \\
		\midrule
		KN5 & 145 & 132 & - & 593 & 169 & - & 141 & 141 & - & 588 & 182 & - \\
		LSTM-LM & 110 & 124 & - & 520 & 165 & - & 105 & 133 & - & 521 & 179 & - \\
		VAE & 112 & 125 & 2 & 317 & 153 & 13 & 106 &  133& 5 & 308 & 164 & 22 \\
		\midrule
		SIVAE-c & 98(\textbf{1.6}) & 121(\textbf{53}) & 5(0.5) & 286(\textbf{2.4}) & 150(\textbf{99}) & 17(1.3) & 94(\textbf{1.6}) & 130(\textbf{56}) & 12(1.0) & 278(\textbf{2.3}) & 161(\textbf{99}) & 29(2.4) \\
		SIVAE-i & \textbf{90}(1.7) & \textbf{119}(60) & 9(1.0) & \textbf{261}(2.6) & \textbf{147}(108) & 24(2.5) & \textbf{89}(1.7) & \textbf{128}(63) & 16(1.9) & \textbf{256}(2.4) & \textbf{158}(104) & 36(5.1) \\
		\bottomrule
	\end{tabular}
	\caption{Language modeling results on testing sets of PTB and wiki90M. For two SIVAE models, the syntactic tree sequence reconstruction scores are shown in parenthesis alongside the sentence reconstruction scores. Lower is better for PPL and NLL. The best results are in bold.}
	\label{tab:lm}
\end{table*}

We conduct our experiments on two datasets: sentence-level Penn Treebank \cite{marcus1993building} with human-constituted parse trees and a 90 million word subset of Wikipedia \cite{gulordava2018colorless} with parsed trees.
When the decoder is too strong, VAE suffers from posterior collapse where the model learns to ignore the latent variable \cite{bowman2016generating}.
To avoid posterior collapse, KL-term annealing and dropping out words during decoding are employed for training in this work.
We also tried an advanced method replacing Gaussian priors with von Mises-Fisher priors \cite{xu2018spherical} to prevent KL collapse, but the results are about the same.

To discover whether the incorporation of syntactic trees is helpful for sentence generation, we compare our two versions of SIVAE with three baselines that do not utilize syntactic information: a 5-gram Kneser-Ney language model (KN5) \cite{heafield2013scalable}, an LSTM language model (LSTM-LM) \cite{sundermeyer2012lstm}, and a standard VAE \cite{bowman2016generating} using an LSTM-based encoder and decoder.
Experimental results of language modeling are evaluated by the reconstruction loss using perplexity and the targeted syntactic evaluation proposed in \citep{marvin2018targeted}.
In section \ref{sec::para}, we show the unsupervised paraphrase generation results.

\paragraph{Datasets}
We use two datasets in this paper.
For sentence-level Penn Treebank (PTB), the syntactic trees are labeled by humans (\textit{i.e.} ``gold-standard'' trees).
For Wikipedia-90M (wiki90M), which does not contain human-generated trees, we first feed the sentences into a state-of-the-art constituency parser \cite{kitaev2018constituency}, and then use the parsed trees as syntactic information for our model.
Further, we replace (low-frequency) words that appear only once in both datasets with the <unk> token.
Statistics about the two datasets are shown in Table \ref{tab:data}.
As we can see, the linearized sequences are much longer than sentences.
The vocabulary of trees sequences is much smaller than the vocabulary of sentences; and golden trees have larger vocabulary than parsed trees.

\paragraph{Settings}
The parameters are fine-tuned on the validation set.
Our implementation of SIVAE uses one-layer bi-directional LSTM architectures for both encoders, and one-layer unidirectional LSTM architectures for both decoders.
The size of hidden units in the LSTM is 600 and the size of word embeddings is 300.
The latent variable size is set to 150 for both sentences and their syntactic trees.
The hidden units size of the MLP in the conditional prior network is 400.
We also tried to use different model sizes for sentences and syntactic trees but the results are about the same and the performance even get worse when the difference of the model sizes is too big.
We use SGD for optimization, with a learning rate of 0.0005.
The batch size is 32 and the number of epochs is 10.
The word dropout rate during decoding is 0.4.
For KL annealing, the initial weights of the KL terms are 0, and then we gradually increase the weights as training progresses, until they reach the KL threshold of 0.8; the rate of this increase is set to 0.5 with respect to the total number of batches.

\subsection{Language Modeling Results}\label{sec::ppl}

We explore two settings for the decoders: standard and inputless.
In the standard setting, the input to the LSTM decoder is the concatenation of the latent representation $\zv$ and the previous ground truth word.
A powerful decoder usually results in good reconstruction in this setting but the model may ignore the latent variable.
In the inputless setting, the decoder purely relies on the latent representations without any use of prior words, so that the model is driven to learn high-quality latent representations of the sentences and syntactic trees.

The language-modeling results, on testing sets evaluated by negative log likelihood (NLL) and perplexity (PPL), are shown in Table \ref{tab:lm}.
SIVAEs outperform all other baselines on both datasets, demonstrating the explicit incorporation of syntactic trees helps with the reconstruction of sentences.
The performance boost on the wiki90M dataset also shows that syntactic trees parsed by a well-developed parser can serve the same function as human-constituted trees, for our model to utilize syntactic information; this underscores how mature parser technology may be leveraged in text generation.
Between the two proposed methods, SIVAE-i is better at reconstructing sentences while SIVAE-c is better at reconstructing syntactic trees.
In the standard setting, VAE performs almost the same as the LSTM language model, possibly because the strong LSTM decoder plays a dominant role when it uses prior words, so the VAE becomes similar to an LSTM language model.
Furthermore, the KL divergence of the proposed models indicate that SIVAE is better at avoiding posterior collapse, so the LSTM sentence decoder can take advantage of the encoded latent variable as well as the previously generated syntactic tree.
In the inputless setting, we see that VAE contains a significantly larger KL term and shows substantial improvement over KN5 and LSTM language models.
SIVAEs further reduces PPL from 317 to 261 on PTB and from 308 to 256 on wiki90M, compared to VAE.

\subsection{Targeted Syntactic Evaluation}\label{sec::grammar}

We adopt targeted syntactic evaluation \citep{marvin2018targeted} to examine whether the proposed methods improve the grammar of generated sentences.
The idea is to assign a higher probability for generating the grammatical sentence than the ungrammatical one, given a pair of sentences that only differ in grammar.
There are three types of sentence pairs used in this work.

\paragraph{Subject-verb agreement (SVA):} Third-person present English verbs need to agree with the number of their subjects. 

For example, \textit{simple SVA}:

(a). The author \underline{laughs}.

(b). *The author \underline{laugh}.

\paragraph{Reflexive anaphoras (RA):} A reflective pronoun such as \textit{himself} needs to agree in number (and gender) with its antecedent. 

For example, \textit{simple RA}:

(a). The senators embarrassed \underline{themselves}.

(b). *The senators embarrassed \underline{herself}.

\paragraph{Negative polarity items (NPI):} Words like \textit{any} and \textit{ever} that can only be used in the scope of negation are negative polarity items.

For example, \textit{simple NPI}:

(a). \underline{No} students have ever lived here.

(b). *\underline{Most} students have ever lived here. 
\newline

In the above examples, we expect the probability of generating (a) to be higher than the probability of generating (b).
However, it is trivial to identify these simple test pairs with simple syntax.
Thus we include complex longer test pairs with greater depth in relative clauses, identifying which requires more understanding of the syntactic structure.

The accuracy per grammar test case of each method is shown in Table \ref{tab:grammar}.
Human scores on these test pairs in \citep{marvin2018targeted} are also shown for reference.
SIVAE outperforms other baselines on grammar testing cases, demonstrating the explicit incorporation of syntactic trees helps with the grammar of generated sentences.
For simple SVA testing pairs, SIVAE-c has a better score than humans.
Even for a difficult grammar test like NPI, our methods still makes significant progress compared to other baselines, whose scores show no syntactic understanding of these sentences.
From Table \ref{tab:grammar}, note that KN5 can only identify simple SVA pairs.
In addition, VAE has similar syntactic performance as a LSTM language model, which verifies the results in reconstruction.
Between the two proposed methods, SIVAE-i makes more grammar mistakes than SIVAE-c, although it has better perplexity in Table \ref{tab:lm}.
This is because SIVAE-c considers the dependency between the sentence prior and the syntactic tree prior, so it can more efficiently prevent the mismatch between two latent variables.
In other words, SIVAE-c learns more robust syntactic representations, but this advantage is not reflected on the reconstruction evaluation.

\begin{table}[t!]
	\centering
	\small
	\begin{tabular}{c|cc|cc|cc}
		\toprule
		\multirow{2}{*}{Model} &
		\multicolumn{2}{c|}{SVA} &
		\multicolumn{2}{c|}{RA} &
		\multicolumn{2}{c}{NPI}\\
		& S & C & S & C & S & C \\
		\midrule
		Humans & 0.96 & 0.85 & 0.96 & 0.87 & 0.98 & 0.81 \\
		\midrule
		KN5 & 0.79 & 0.50 & 0.50 & 0.50 & 0.50 & 0.50 \\
		LSTM-LM & 0.94 & 0.56 & 0.83 & 0.55 & 0.50 & 0.50 \\
		VAE & 0.94 & 0.57 & 0.84 & 0.57 & 0.51 & 0.50 \\
		\midrule
		SIVAE-c & \textbf{0.97} & \textbf{0.75} & \textbf{0.89} & \textbf{0.64} & \textbf{0.57} & \textbf{0.52} \\
		SIVAE-i & 0.95 & 0.71 & 0.88 & 0.63 & 0.56 & 0.52 \\
		\bottomrule
	\end{tabular}
	\caption{Accuracy of targeted syntactic evaluation for each grammar test case. S and C denote simple testing pairs and complex testing pairs. The total number of test sentences is 44800. Models are trained on wiki90M. The best results are in bold.}
	\label{tab:grammar}
\end{table}

\subsection{Unsupervised Paraphrasing Results}\label{sec::para}
The proposed method is used for generating paraphrases by implicitly selecting (SIVAE-c) or explicitly changing (SIVAE-i) the syntactic tree templates.
Our model is not trained on a paraphrase corpora, which makes it a purely unsupervised paraphrasing network.

\begin{table*}[t!]
	\centering  
	\small
	\begin{tabular}{p{2.2in}p{3.8in}} \toprule
		Template & Paraphrase \\ 
		\midrule
		original & the discovery of dinosaurs has long been accompanied by a legend .  \\
		( SBARQ ( NP ) ( VP ) ( , ) ( SQ ) ( ? ) ) & the discovery of dinosaurs has been a legend , is it ? \\
		( S ( `` ) ( NP )  ( VP ) ( '' ) ( NP ) ( VP ) ( . ) )& `` the discovery of dinosaurs is a legend '' he said .  \\
		( S ( VP ) ( , ) ( NP )  ( . ) ) & \textcolor{blue}{having been accompanied , the unk lengend . }\\
		\midrule
		original & in 1987 a clock tower and a fountain were erected at council unk monument .   \\
		( S ( PP ) ( PP ) ( NP ) ( VP ) ( . ) ) & in 1987 at council a fountain was erected . \\ 
		( S ( VP ) ( NP ) ( CC ) ( NP ) ( PP ) ( . )  )& build a clock and a fountain at council unk unk . \\
		( S ( NP ) ( ; ) ( S ) ( PP ) ( . ) ) & \textcolor{blue}{a clock p ; he shops everything on the fountain at unk unk .} \\\bottomrule
	\end{tabular}
	\caption{Examples of syntactically controlled paraphrases generated by SIVAE-i. We show two successful and one failed (in blue) generations with different templates for each input sentence.} \label{tab:para2}
\end{table*}

\paragraph{Syntactically Controlled Paraphrasing}
SIVAE-i as the syntactically controlled paraphrasing network is trained on sentences and their simplified syntactic sequences of PTB and wiki90M dataset.
Table \ref{tab:para2} shows some example paraphrases generated by SIVAE-i using different syntactic templates.
We see that SIVAE-i has the ability to syntactically control the generated sentences that conform to the target syntactic template.
The examples are well-formed, semantically sensible, and grammatically correct sentences that also preserve semantics of the original sentences.
However, the model can generate nonsensical outputs, like the failed cases in Table \ref{tab:para2}, when the target template mismatches the input sentence.

\begin{table}
		\centering
		\def\arraystretch{1.0}
		\setlength{\tabcolsep}{3pt}
		\small
			\begin{tabular} {c p{2.5in}}
				\toprule
				\textbf{Ori} & the new york times has been one of the best selling newspapers in america . \\ 
				\midrule
				\textbf{Gen1} & the new york times also has been used as american best selling newspaper . \\
				\midrule
				\textbf{Gen2} & the new york times also has been used as a `` unk '' that sells in america . \\
				\midrule
				\textbf{Gen3} &  the new york times also has been used as the best `` unk '' selling in america . \\
				\bottomrule
			\end{tabular}
		\caption{An example of paraphrases generated by SIVAE-c.}
		\label{tab:para1} 
\end{table}

\paragraph{Paraphrasing with Different Tempera}
We further perform paraphrasing using SIVAE-c with different tempera.
Table \ref{tab:para1} shows example paraphrases generated by SIVAE-i.
We see that SIVAE-c can generate grammatical sentences that are relevant to the original sentence.
However, the generated paraphrases are very similar, indicating that the variance of the conditional prior network is small.
In other words, given a sentence latent representation, the range for SIVAE-c selecting a possible syntactic tree representation is small, so it tends to generate similar paraphrases.

\paragraph{Qualitative Human Evaluation}
We adopt similar human evaluation metrics as in \cite{gupta2018deep} for generated paraphrases.
For 20 original sentences, we collect 5 paraphrases for each sentence (100 in total) generated by SIVAE-c or SIVAE-i using $5$ different syntactic templates.
The models are trained on PTB and wiki90M.
Three aspects are verified in human evaluation: \textit{Relevance} with the original sentence, \textit{Readability} w.r.t. the syntax of generated sentences, and \textit{Diversity} of different generations for the same original sentence.
Three human evaluators assign a score on a scale of 1-5 (higher is better) for each aspect per generation. 

The human evaluation results for unsupervised paraphrase generation using standard VAE, SIVAE-i and SIVAE-c are shown in Table \ref{tab:human}.
SIVAE-c has the best scores and standard VAE has the worst scores at the readability of generated sentences, which further verifies that syntactic information is helpful for sentence generation.
Paraphrases generated by SIVAE-i are more diverse under different syntactic templates, compared to SIVAE-c and standard VAE.
All three models show better paraphrasing performance on the wiki90M dataset.

\begin{table}[t!]
	\centering
	\small
	\begin{tabular}{c|ccc|ccc}
		\toprule
		\multirow{2}{*}{Model} &
		\multicolumn{3}{c|}{PTB} &
		\multicolumn{3}{c}{wiki90M} \\
		& Rele & Read & Div & Rele & Read & Div \\
		\midrule
		VAE & 2.63 & 3.07 & 2.77 & 3.03 & 3.20 & 2.60 \\
		\midrule		
		SIVAE-c & 2.93 & 3.47 & 2.80 & 3.27 & 3.67 & 2.73 \\
		SIVAE-i & 3.00 & 3.30 & 3.13 & 3.37 & 3.53 & 3.20 \\
		\bottomrule
	\end{tabular}
	\caption{Human evaluation results on Relevance, Readability, and Diversity of generated paraphrases.}
	\label{tab:human}
\end{table}

\subsection{Continuity of Latent Spaces}
We further test the continuity of latent spaces in our model.
Two vectors $\zv_A$ and $\zv_B$ are randomly sampled from the sentence latent space of SIVAE-c.
Table \ref{tab:continuity} shows generated sentences based on intermediate points between $\zv_A$ and $\zv_B$.
We see the transitions are smooth and the generations are grammatical, verifying the continuity of the sentence latent space.
The syntactic structure remains consistent in neighborhoods along the path, indicating the continuity in the syntactic tree latent space.

\begin{table}
	\def\arraystretch{1.0}
	\setlength{\tabcolsep}{3pt}
	\small
	\begin{tabular} {c p{2.8in}}
		\toprule
		\textbf{A} & in january 2014 , the unk announced that one player would be one of the first two heroes . \\ 
		\midrule
		$\bullet$ & in january 2014 , he was one of the first two players to be the most successful . \\
		\midrule
		$\bullet$ & until the end of the first half of the series , he has played the most reported time . \\
		\midrule
		$\bullet$ &  until the end of world war i , he was the first player in the united states .  \\
		\midrule
		$\bullet$ & there are also a number of other members in the american war association .  \\
		\midrule
		\textbf{B} & there are also a number of other american advances , such as the unk unk of the american association .  \\
		\bottomrule
	\end{tabular}
	\caption{Intermediate sentences are generated between two random points in the latent space of SIVAE-c.} 
	\label{tab:continuity}
\end{table}

\section{Conclusion}
We present SIVAE, a novel syntax-infused variation autoencoder architecture for text generation,  leveraging constituency parse tree structure as the linguistic prior to generate more fluent and grammatical sentences.
The new lower bound objective accommodates two latent spaces, for jointly encoding and decoding sentences and their syntactic trees.
The first version of SIVAE exploits the dependencies between two latent spaces, while the second version enables syntactically controlled sentence generation by assuming the two priors are independent.
Experimental results demonstrate the incorporation of syntactic trees is helpful for reconstruction and grammar of generated sentences.
In addition, SIVAE can perform unsupervised paraphrasing with different syntactic tree templates.

\section*{Acknowledgments}
This research was supported in part by DARPA, DOE, NIH, ONR and NSF.

We thank Kazi Shefaet Rahman,  Ozan Irsoy, Igor Malioutov and other people in the Bloomberg NLP platform team for their feedback on the initial idea of the work. We thank the ACL reviewers for their helpful feedback. This work also benefitted from discussions with Tao Lei and Lili Yu.

\bibliographystyle{acl_natbib}
\bibliography{ref}



\end{document}